\documentclass[letterpaper, 10 pt, conference]{ieeeconf}  

\IEEEoverridecommandlockouts                              

\usepackage{graphics} 
\usepackage{epsfig} 
\usepackage{mathptmx} 
\usepackage{times} 
\usepackage{amsmath} 
\usepackage{amssymb}  
\usepackage[normalem]{ulem}
\usepackage{cite}

\usepackage{mathtools, nccmath}
\DeclarePairedDelimiter{\nint}\lfloor\rceil

\usepackage{color}
\usepackage{soul}
\usepackage{multirow}
\usepackage{rotating}
\usepackage{wrapfig}
\definecolor{turquoise}{cmyk}{0.65,0,0.1,0.1}
\definecolor{purple}{rgb}{0.65,0,0.65}
\definecolor{darkgreen}{rgb}{0.0, 0.5, 0.0}
\definecolor{darkred}{rgb}{0.5, 0.0, 0.0}
\definecolor{darkblue}{rgb}{0.0, 0.0, 0.5}
\definecolor{blue}{rgb}{0.0, 0.0, 1.0}
\definecolor{orange}{rgb}{1.0, 0.5, 0.0}

\newcommand{\hide}[1]{{}}

\title{LeRoP: A Learning-Based Modular Robot Photography Framework}

\author{Hao Kang$^{1}$, Jianming Zhang$^{2}$, Haoxiang Li$^{3}$, Zhe Lin$^{2}$, TJ Rhodes$^{2}$, Bedrich Benes$^{1}$
\thanks{$^{1}$Hao Kang and Bedrich Benes are with the Purdue University, 610 Purdue Mall, West Lafayette, IN 47901, USA {\tt\small \{kang235, bbenes\} @purdue.edu}}
\thanks{$^{2}$Jianming Zhang, Zhe Lin and TJ Rhodes are with the Adobe Research, 345 Park Ave., San Jose, CA 95110, USA {\tt\small 
\{jianmzha, zlin, trhodes\}@adobe.com}}
\thanks{$^{3}$Haoxiang Li is with the AIBee, 350 Cambridge Avenue, Suite 250, Palo Alto, CA 94306, USA {\tt\small 
hxli@aibee.com}}
}

\begin{document}

\maketitle
\thispagestyle{empty}
\pagestyle{empty}

\begin{abstract}
We introduce a novel framework for automatic capturing of human portraits. The framework allows the robot to follow a person to the desired location using a Person Re-identification model. When composing is activated, the robot attempts to adjust its position to form the view that can best match the given template image, and finally takes a photograph. A template image can be predicted dynamically using an off-the-shelf photo evaluation model by the framework, or selected manually from a pre-defined set by the user. The template matching-based view adjustment is driven by a deep reinforcement learning network. Our framework lies on top of the Robot Operating System (ROS). The framework is designed to be modular so that all the models can be flexibly replaced based on needs. We show our framework on a variety of examples. In particular, we tested it in three indoor scenes and used it to take 20 photos of each scene: ten for the pre-defined template, ten for the dynamically generated ones. The average number of adjustment was $11.20$ for pre-defined templates and $12.76$ for dynamically generated ones; the average time spent was $22.11$ and $24.10$ seconds respectively.
\end{abstract}

\section{Introduction}
Getting a good picture requires getting into a fine location, proper light, and right timing.
While getting a great picture requires talent and artistic preparation, there is a possible way of getting a well-taken image automatically.
For example, a system could follow the rule of thirds, empty area, keeping eyes in sight, and other commonly-accepted conventions~\cite{Grill1990}.

Some previous work has been done in this direction. For example, Kim et al.~\cite{Kim2010} introduced a robot that can move and capture photographs according to composition lines of the human target. Robot Photographer of Luke~\cite{Zabarauskas2014} can randomly walk in an unstructured environment, and take photographs of humans basing on heuristic composition rules. To get a good picture, we need to move the robot to a certain location. At the same time, a good sense of composition needs to be taught to the robot. However, the robustness of geometric-based robot motion planning relies heavily on sensor accuracy. Moreover, traditional rule-based designs are usually rigid in handling various composition situations. 

Therefore, we propose a flexible learning-based view adjustment framework for taking indoor portraits that we call: LeRoP. The objective of our work is to create a framework that can train a robot to automatically move and capture the best view of a person. We implemented the LeRoP utilizing a photo evaluation model to propose good views, and using a Deep Reinforcement Learning (DRL) model to adjust the robot position and orientation towards the best view to capture. Additionally, the framework is interactive to the user. For example, our robot with the framework can be triggered to follow the target to a photograph location before searching for the best view. We ingeniously utilized a $360$ camera as a supplement to the main camera (for photography). The omindirectional view helps avoid repetitive rotations in tracking and selects views for the DRL template matching procedure. The modular design allows the photo evaluation aesthetic model to be swapped flexibly basing on photo style preferences. The DRL model can adapt to different view adjustment methods by simply re-training the network with a suitable reward function and apply to different hardware settings.

\begin{figure}[!hbt]
\centering
\includegraphics[width=\linewidth]{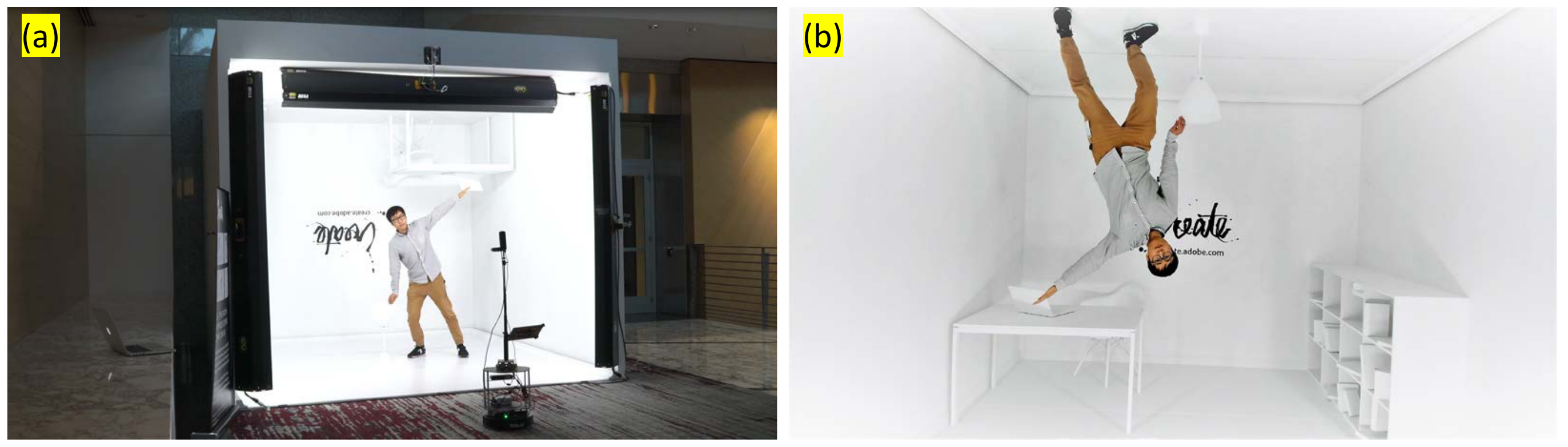}
\caption{Our LeRoP robot at work. Photo (a) is the third-person view of the working scene. Photo (b) is the final capture (the photo has been rotated by $180^{\circ}$ for better visualization).}
\label{fig:teaser}
\end{figure}

An example in Figure~\ref{fig:teaser} shows our framework at work. The user first selects \textit{Tracking Mode} to lead the robot to a user desired location. The user then selects \textit{Composing Mode} to start the autonomous composing process. Once \textit{Composing Mode} is activated, the robot observes the scene with its $360^{\circ}$ camera for hunting the best view (template). Once the template is detected, it moves into a location matches the template that allows for taking a high-resolution portrait with the second camera.

We tested it on a robot system built on \textit{Turtlebot}~shown in Figure~\ref{fig:sohw}, and our robot can (a) interact with the user, (b) identify and follow the user, (c) propose well-composed template dynamically or use supplemental pre-defined template, and (d) adjust position to match the template and capture the portrait.  

We claim the following contributions:
\begin{enumerate}
\item A template matching based DRL solution with its synthetic virtual training environment for robot view adjustment.
\item A method to utilize an omnidirectional camera to support tracking and DRL view selection.
\item An interactive modular robot framework design that supports automatically capture high quality human portraits.
\end{enumerate}
\section{Related Work}
Autonomous cameras for both virtual and real photography have long been explored. We refer the reader to the review~\cite{Chen2014} that summarizes autonomous cameras from the viewpoint of camera planning, controlling, and target selecting. Galvane et al.~\cite{Galvane2013, Galvane2014, Galvane2015} also provided several studies on automatic cinematography and editing for virtual environments. 

One of the earliest robotic cameras was introduced in~\cite{Pinhanez1995,Pinhanez1997}. This camera allows intelligent framing of subjects and objects in a TV studio upon verbal request and script information. Byers et al.~\cite{Byers2003} developed probably the first robot photographer that can navigate with collision avoidance and frame image using face detection and predefined rules. They expanded their work by discussing their observations, experiences, and further plans with the robot in follow up studies~\cite{Smart2003,Byers2004}. Four principles from photography perspective~\cite{Grill1990} were applied to their system and to many later robot photographer systems. In particular they implemented the rule of thirds, the empty-space rule, the no-middle rule, and the edge rule. 

The follow-up work~\cite{Ahn2006} extended the work of Byers et al.~\cite{Byers2003,Smart2003,Byers2004} by making the robot photographer interactive with users. A framework introduced in~\cite{Zabarauskas2014} described RGB-D data-based solutions with Microsoft Kinect and the capability of detecting direction via human voice recognition was added in~\cite{Kim2010}. Campbell et al.~\cite{Campbell2005} introduced a mobile robot system that can automate group-picture-framing by applying optical flow and motion parallax techniques. The photo composing in most of these studies rely on heuristic composition rules or similar techniques. Such settings usually do not adequately account for the correlation between the foreground and the background, as well as the light and color effects of the entire image. They also lack a generalizable framework for varied application scenarios. 

Recent studies use UAV technology for drone photography trajectory planning ~\cite{Gebhardt2016, Richter2016, Roberts2016} and drone photography system redesign~\cite{Kang2017, Lan2017}. Although these studies can make it easier for users to obtain high-quality photos, the methods usually only provide an auxiliary semi-automated photo taking process. Users still need to use subjective composition principles to get satisfactory photos.

One crucial part in autonomous 
camera photography is view selection (\emph{i.e.}~framing). Besides the composition principles~\cite{Grill1990} that are widely used on robot photography ~\cite{Gadde2011,Gooch2001,Cavalcanti2006,Banerjee2007} for view selection, there are many recent efforts  
in photo quality evaluation and aesthetics analysis using computational methods~\cite{Datta2006,Dhar2011,Ke2006,Luo2011,Nishiyama2011}. These methods use machine learning algorithms to learn aesthetic models for image aesthetic evaluation. 
The recent advances in deep learning further elevate the research in this direction. Large datasets containing photo rankings basing on aesthetics and attributes are curated ~\cite{Murray2012,Kong2016,Wei2018} and they allow training deep neural networks to predict photo aesthetics levels and composition quality~\cite{Lu2015,Lu2014,Marchesotti2015,Mai2016,Kang2014,Wei2018}. Compared with aesthetic evaluation methods based on pre-set rules, these data-driven approaches can handle more general and complex scenes where these rules cannot simply be applied. Therefore, we embrace the deep learning based aesthetic models for view selection in our system.

The success in Altari 2600 video games \cite{Mnih2015} and AlphaGo \cite{Silver2016} showed the power of DRL in solving decision-making problems. DRL also enables the control policy end-to-end learning for virtual agents \cite{Peng2017, Hodgins2017}, and real robots \cite{Gu2017, Hwangbo2017}. Our study can be simplified as a visual servoing problem of robot special navigation. Several DRL driven visual servoing \cite{Levine2016, Levine2018} and navigation \cite{Mirowski2016, Zhu2017} scenarios were well studied to allow their autonomous agents to interact with the environments. The previous research proved the capability and aptness of using DRL to learn optimal behaviors effectively and efficiently.

\section{System Overview}
We designed the framework on the robot to: interact with the user, track the user to the desired location, and adjust the position that can take a well-composed portrait. The framework will be systematically discussed from both the hardware and software perspectives.

\subsection{Hardware}
\begin{figure}[!hbt]
\centering
\includegraphics[width=\linewidth]{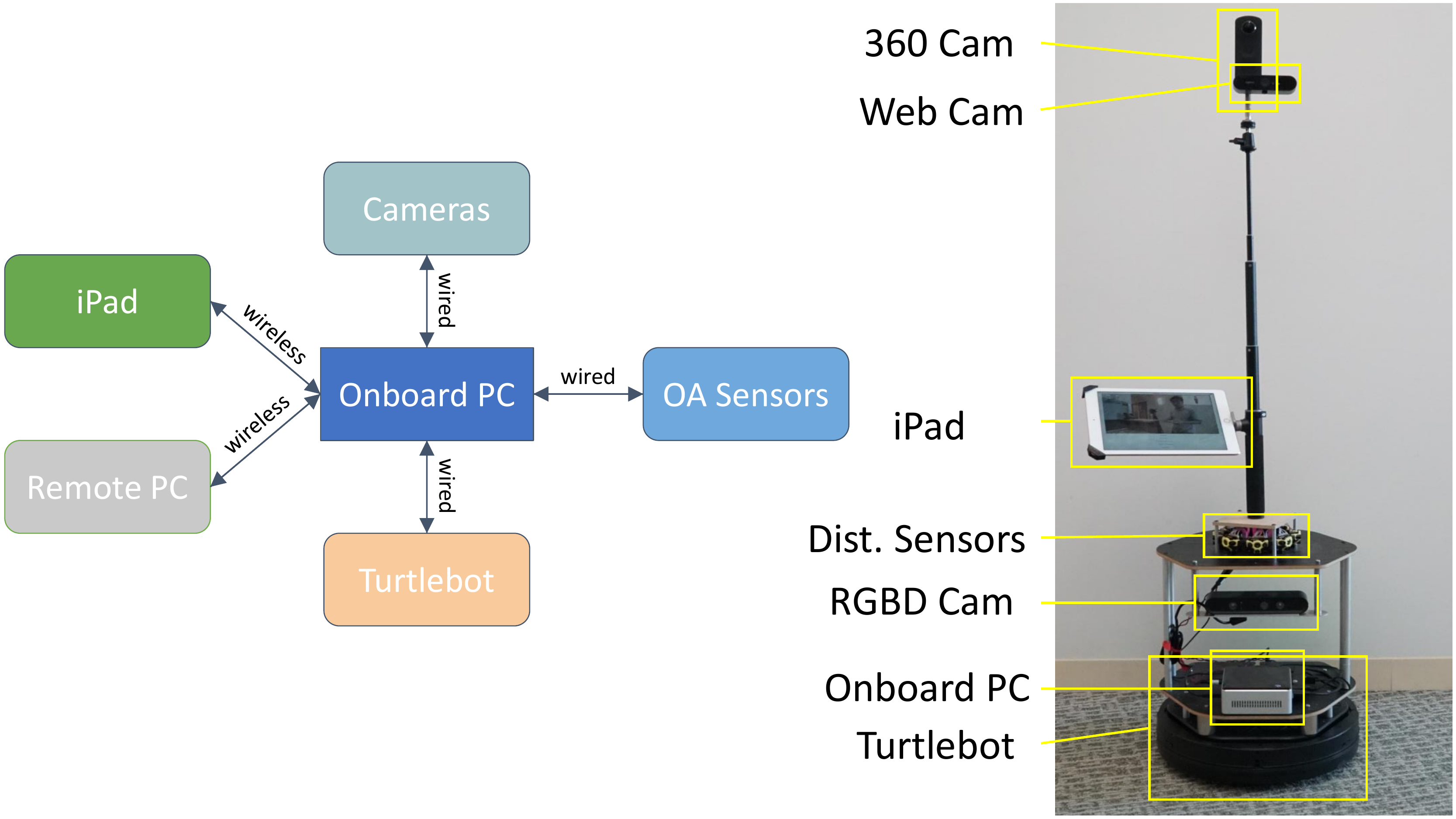}
\caption{Hardware system overview. }
\label{fig:sohw}
\end{figure}

The hardware of the entire framework consists of eight major components shown in Figure~\ref{fig:sohw} and the devices with corresponding models are listed in Table~\ref{tab:sohw}. 





\begin{table}[!hbt]
\caption{The hardware configuration.}
\resizebox{\columnwidth}{!}{%
\begin{tabular}{|c|c|}
\hline
\textbf{Device} & \textbf{Model}\\ \hline
360 Camera &  Ricoh Theta S\\ \hline
Web Camera &  Logitech Brio 4K\\ \hline
RGB-D Camera &  Orbbec Astra\\ \hline
Tablet Computer & iPad\\ \hline
Distance Sensors & TeraRanger Tower\\ \hline
Mobile Base & Yujin Turtlebot 2\\ \hline
 & CPU: Intel Core i7-5557U@3.1GHz\\
Onboard PC & RAM: 8GB\\
 & GPU: Intel Iris Graphics 6100\\ \hline
 & CPU: Intel Xeon E5-2603v3@1.60GHz\\
Remote PC & RAM: 16GB \\
 & GPU: Nvidia Geforce GTX 1080\\ \hline
\end{tabular}
}
\label{tab:sohw}
\end{table}

\subsection{Software}
\begin{figure}[!hbt]
\centering
\includegraphics[width=\linewidth]{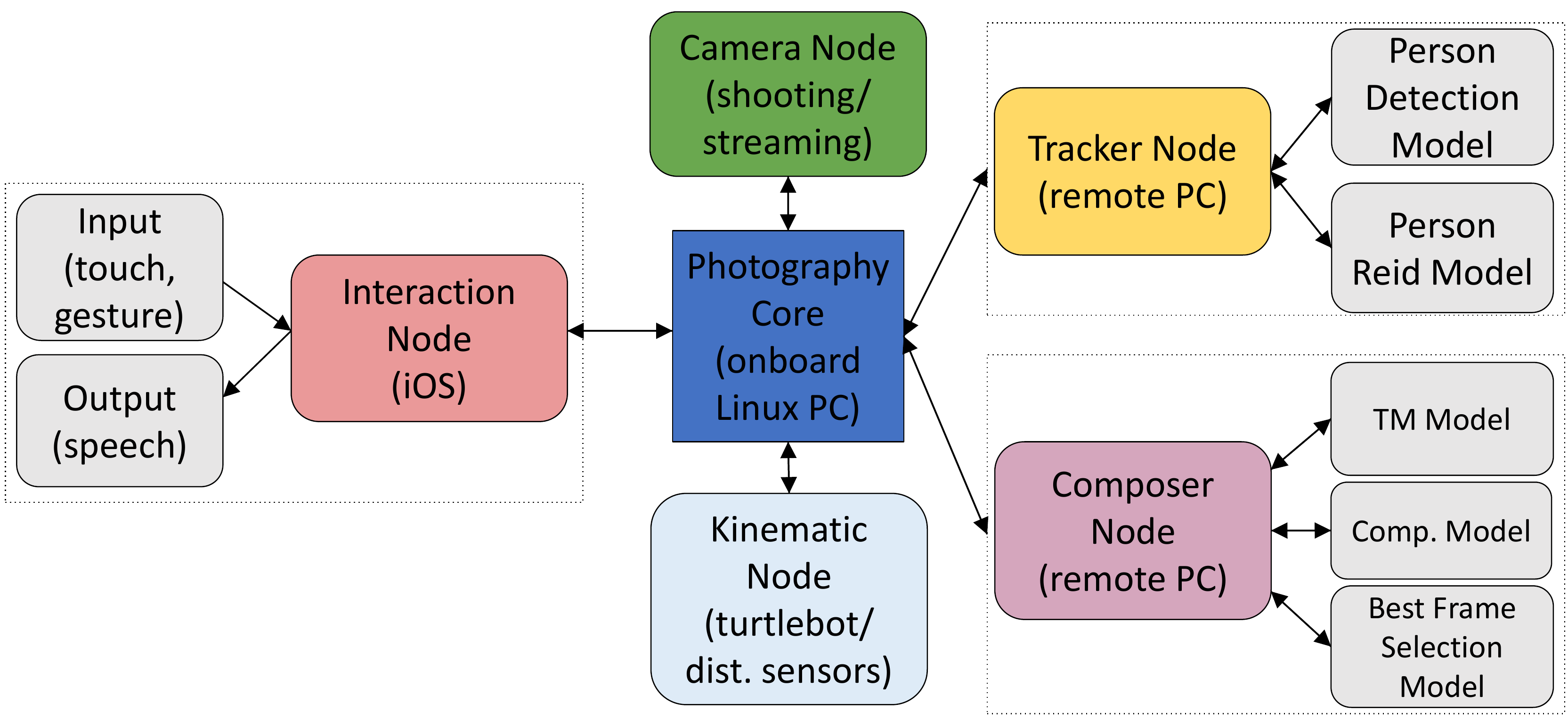}
\caption{Software framework overview.}
\label{fig:sosw}
\end{figure}

The software framework is built on top of \textit{Robot Operating System} (ROS)~\cite{Quigley2009}, and the architecture is shown in Figure~\ref{fig:sosw}. 

The \textit{Core Node} runs on the on-board PC, and it is in charge of communication between nodes on controlling different hardware and software components. 

The \textit{Kinematic Node} controls the linear and angular motion of the robot, and avoids collision with obstacles. 

The \textit{Camera Node} provides the framework with vision ability such as real-time video streaming and photo shooting. The Kinematic and Camera nodes both reside on the on-board PC. The vision is essential to the robot photographer, relating to two major modes: \textit{Following Mode} and \textit{Photographing Mode}. The two modes can be activated and switched through the Interaction Node which is presented as an iPad application. 

The application has a \textit{Graphical User Interface} (GUI) that takes touch gestures and human poses as input, and gives graphical results and voice prompts as output (Section~\ref{sec:interaction}). 
The two modes are implemented with two separate nodes. The \textit{Tracker Node} (Section~\ref{sec:tracker}) is deployed to the remote PC, which analyzes real-time video stream using Person Detection and Re-Identification Neural Network models to identify and follow the user. The tracking is made with the help of wide-view panorama images provided by the 360 camera and the depth information supported by the RGB-D camera. The \textit{Composer Node} also resides on the remote PC. Composer Node (Section~\ref{sec:composer}) utilizes a Deep Neural Network (DNN) Composition model to determine: the best target view (template), adjust the robot towards the target view with a Deep Reinforcement Learning (DRL) Template Matching Model, and finally shoot the target and select the best photo from candidates with a DNN Best Frame Selection model.  
\section{Implementation}\label{sec:sw}

\subsection{The Tracker}
\label{sec:tracker}
\begin{figure}[hbt!]
\centering
\includegraphics[width=\linewidth]{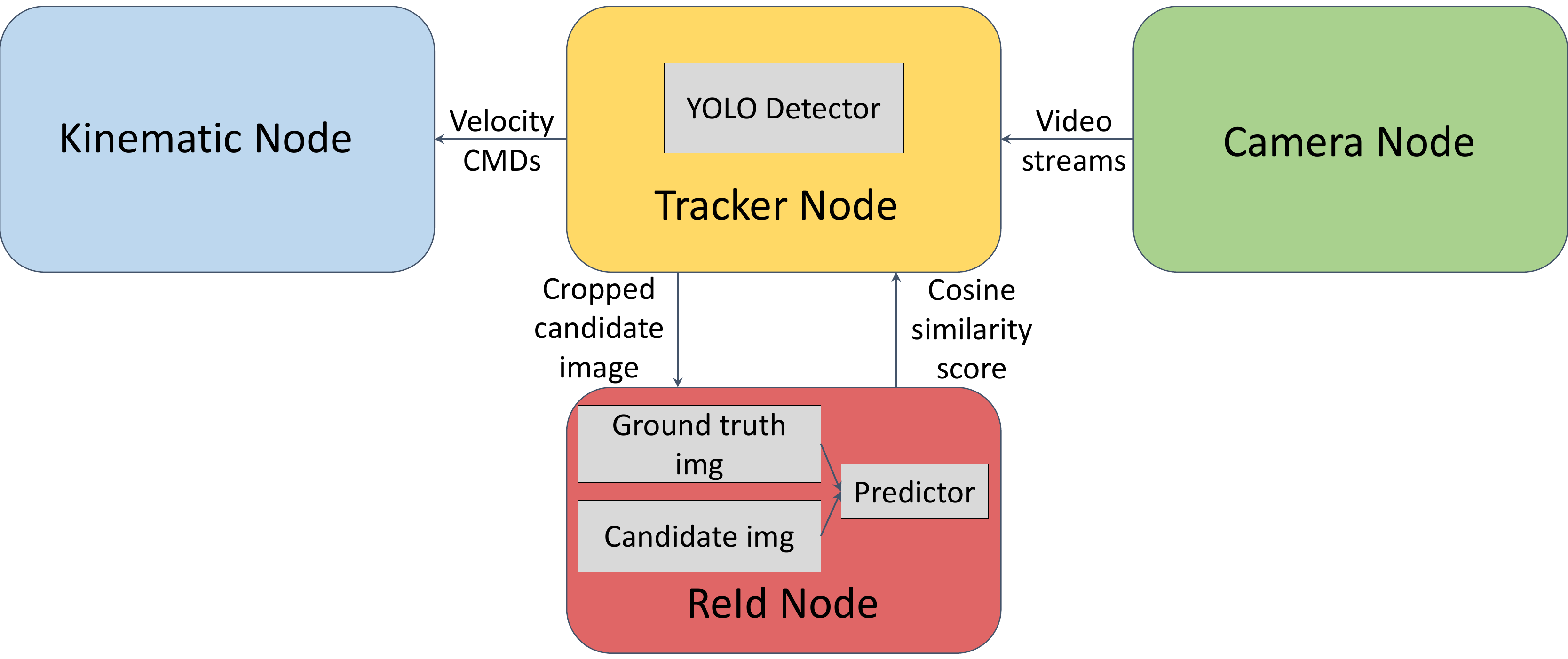}
\caption{The tracker module architecture.}
\label{fig:tracker_block}
\end{figure}

The \textit{Tracker Module} is responsible for the spatial tracking of the user. It allows the robot to generate instructions for the \textit{Kinematic Node} to follow the user. The architecture of the \textit{Tracker Module} is presented in Figure~\ref{fig:tracker_block}.

Our tracker module utilizes the images from the 360 and RGB-D cameras. The panorama image provides a $360^{\circ}$ view for omni-directionally searching and identifying the user around the robot. When the user is located, the robot rotates until the user is centered in the RGB-D camera view. The depth information is used to retrieve the distance between the robot and the target. The distance value determines the linear tracking velocity of the robot. A velocity smoother~\cite{Yujin2018} is used to control the robot's acceleration. Tracking is not activated when the user is within "operating zone", which is within ${0.5}$ meters around the robot. When obstacles less than $0.5$ meters are detected by OA sensors, the linear velocity of the robot decreases to zero. When the tracking target is missing from the view, the robot stops and waits for the target to appear or the tracker to be reset. 


The \textit{Tracker Node} uses YOLO~\cite{Redmon2016} to generate candidate person bounding boxes for the input panoramic images. The candidate person images are cropped out and broadcast to the \textit{ReId Node} that uses the person ReId model from \cite{Hermans2017} to compare the reference person image with each candidate person image, and predicts cosine similarity scores. The candidate person most similar to the ground truth person is considered to be the target with a goal drop threshold ($0.80$). The reference person image is initially set when the tracker is activated by the user, and continuously updated when the score of a candidate is higher than the threshold ($0.95$). The processing time for the tracker node is about $0.04$ seconds for a $1440\times360$ input image. The velocity commands generated by the \textit{Tracker Node} are broadcast and received by the \textit{Kinematic Node} for robot motions.


\begin{figure}[!hbt]
\centering
\includegraphics[width=0.99\linewidth]{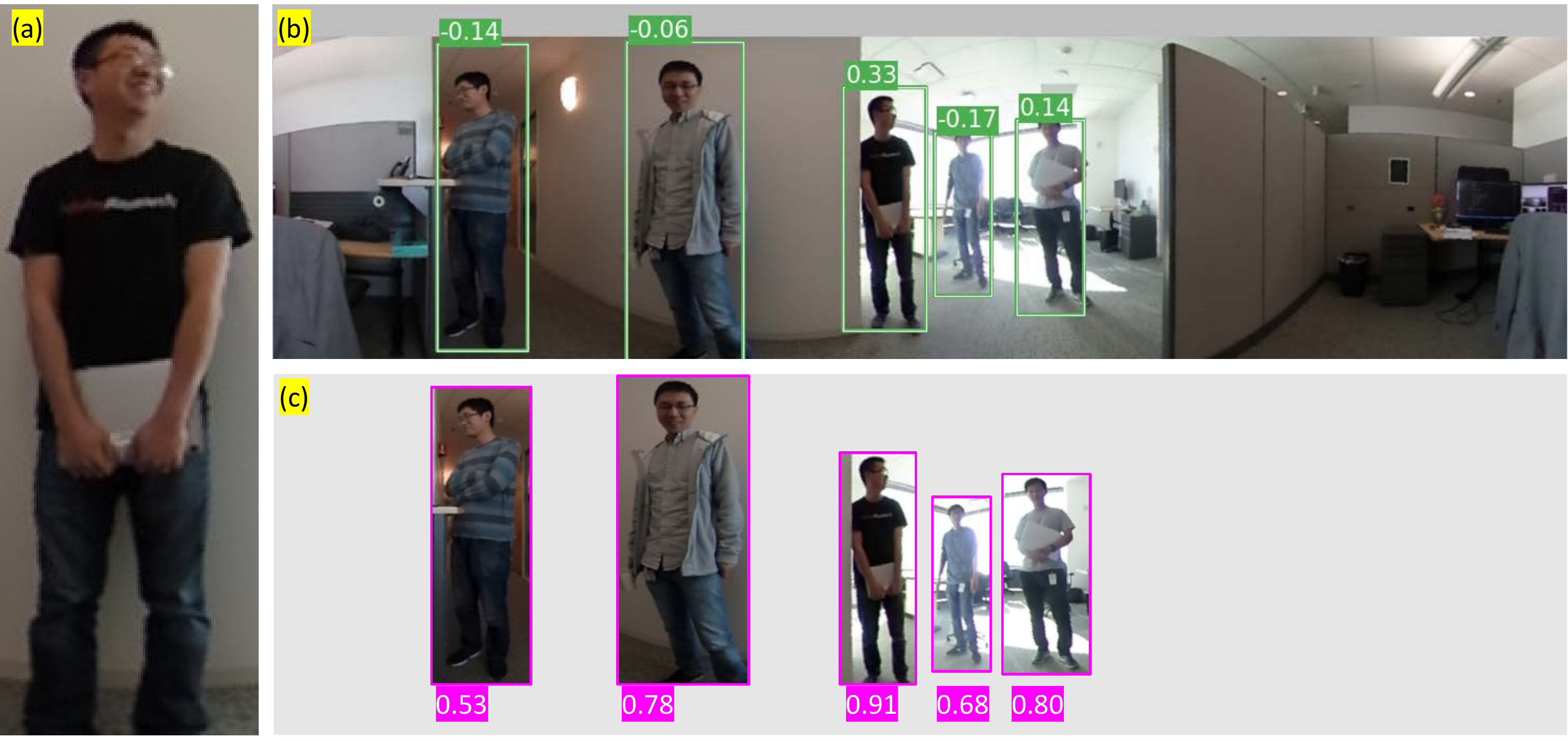}
\caption{\textit{Person ReId} results. The ground truth person image in (a) is the query person to track. Part (b) demonstrates the proposed bounding boxes of candidates, and their similarity scores using the joint ReId model~\cite{Xiao2017}. The reference person images in (c) are generated with YOLO~\cite{Redmon2016}. The similarity scores in (c) are predicted with the triplet-loss ReId model~\cite{Hermans2017}. Both methods can provide correct result. Note that the scores in (b) and (c) are not normalized to the same scale.}
\label{fig:reid}
\end{figure}

\subsection{The Composer}\label{sec:composer}
\begin{figure}[!hbt]
\centering
\includegraphics[width=0.99\linewidth]{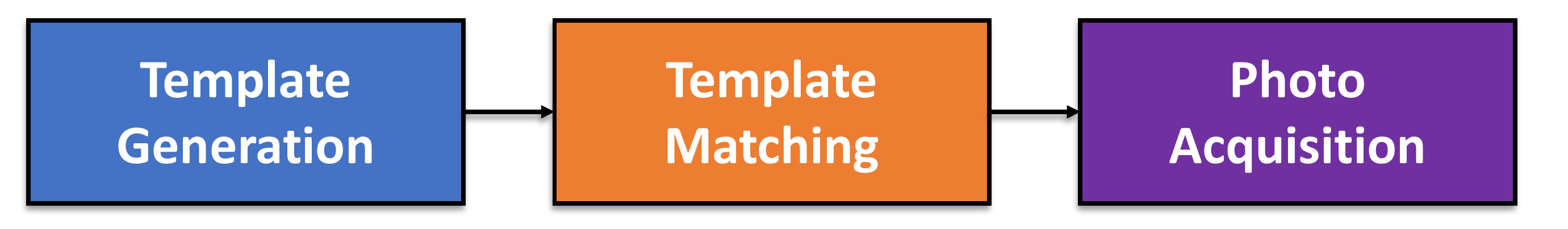}
\caption{The pipeline of photo composing.}
\label{fig:photographerpipline}
\end{figure}

After the robot tracks the user to a desired photography location, photo composing can be activated to automatically adjust the robot position and take photos for the user based on (a) the static pre-defined templates (Figure~\ref{fig:pre-template}a) or (b) dynamically proposed well-composed views (Figure~\ref{fig:composition}d). The pipeline for the photo composing process is described in Figure~\ref{fig:photographerpipline}.

\subsubsection{Template Generation}
A template contains information that is used to guide the robot composing a user satisfied photo. We use three input data from a template: (a) the location of the person in the photo, (b) the size of the person in the photo, and (c) the pose of the person in the photo. The template can be chosen manually or generated automatically. 

\begin{figure}[!hbt]
\centering
\includegraphics[width=0.6\linewidth]{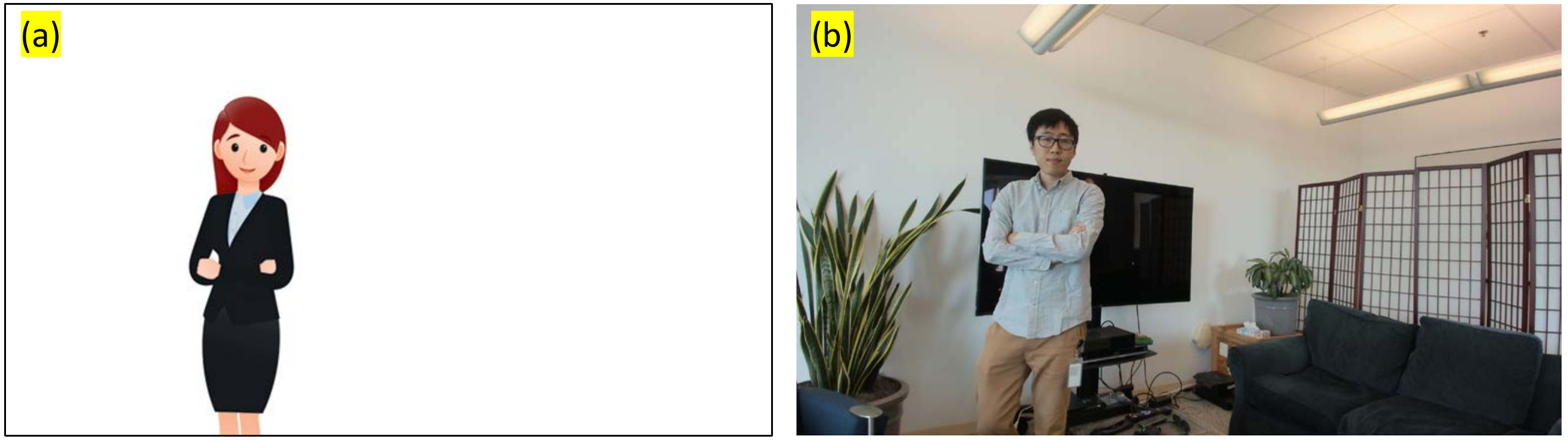}
\caption{Pre-defined templates can be used for photo composing. (a) demonstrates an example of a pre-defined template. (b) shows the final capture with the template from (a).}
\label{fig:pre-template}
\end{figure}

There is a set of pre-defined templates varying in location, size, and pose for the user to manually pick from the system. Figure~\ref{fig:pre-template}a demonstrates a pre-defined template with a cartoon avatar. With this template set, the robot moves around to compose the final photo that matches the template (Section~\ref{sec:matching}) for the user (Figure~\ref{fig:pre-template}b). The pre-defined template is not necessarily a cartoon image; the system supports any single person photo with proper aspect ratio.  

\begin{figure*}[hbt]
\centering
\includegraphics[width=\linewidth]{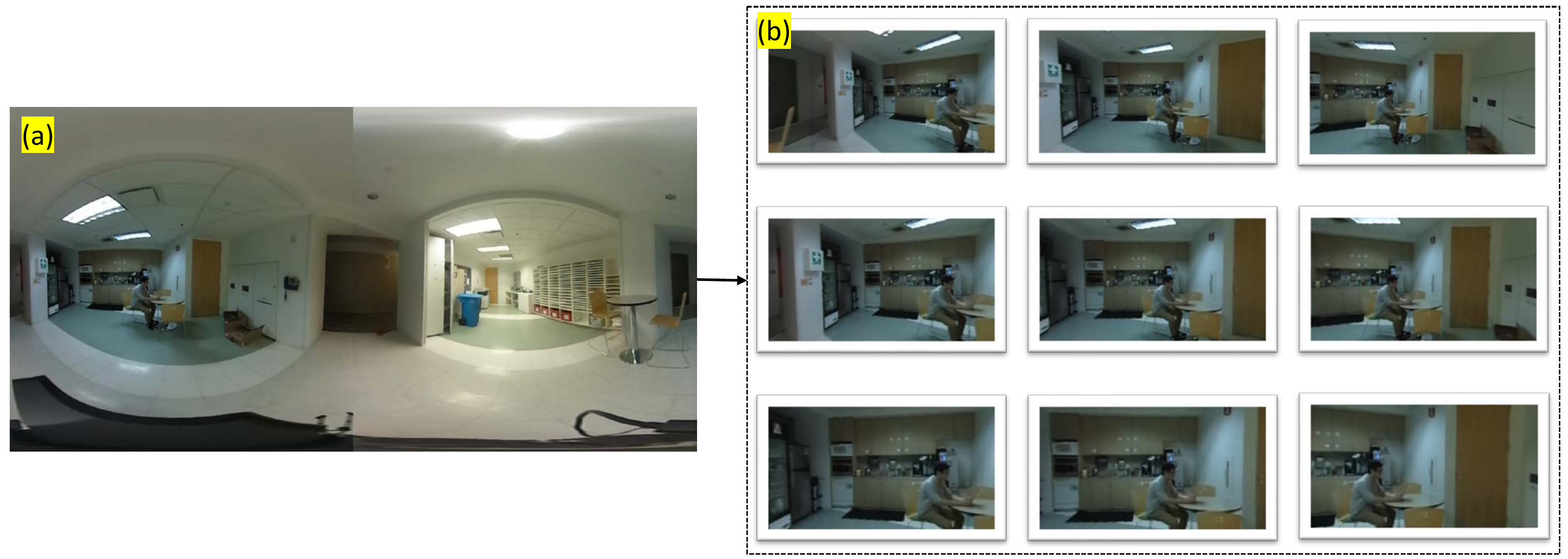}
\caption{The panorama photo in (a) is cropped and de-warped to form a collection of candidate templates in (b). }
\label{fig:unwarp}
\end{figure*}

The system also provides a dynamic template generation solution for photo composing. The novel modular solution enables autonomous photographing with the robot. The solution requires the panorama photos from the 360 camera and the final capture with the high-quality webcam. An example panorama photo is shown in Figure~\ref{fig:unwarp}a. The panorama photo is cropped and remapped with the method described in~\cite{Mo2018} to form a collection of candidate templates (Figure~\ref{fig:unwarp}b). The candidate templates are generated with different levels of distance and yaw angles. Each candidate template is guaranteed to contain the person target. The remapping de-warps the images with the camera parameters of the webcam to make sure the view is reachable from webcam. 

\begin{figure}[!hbt]
\centering
\includegraphics[width=0.99\linewidth]{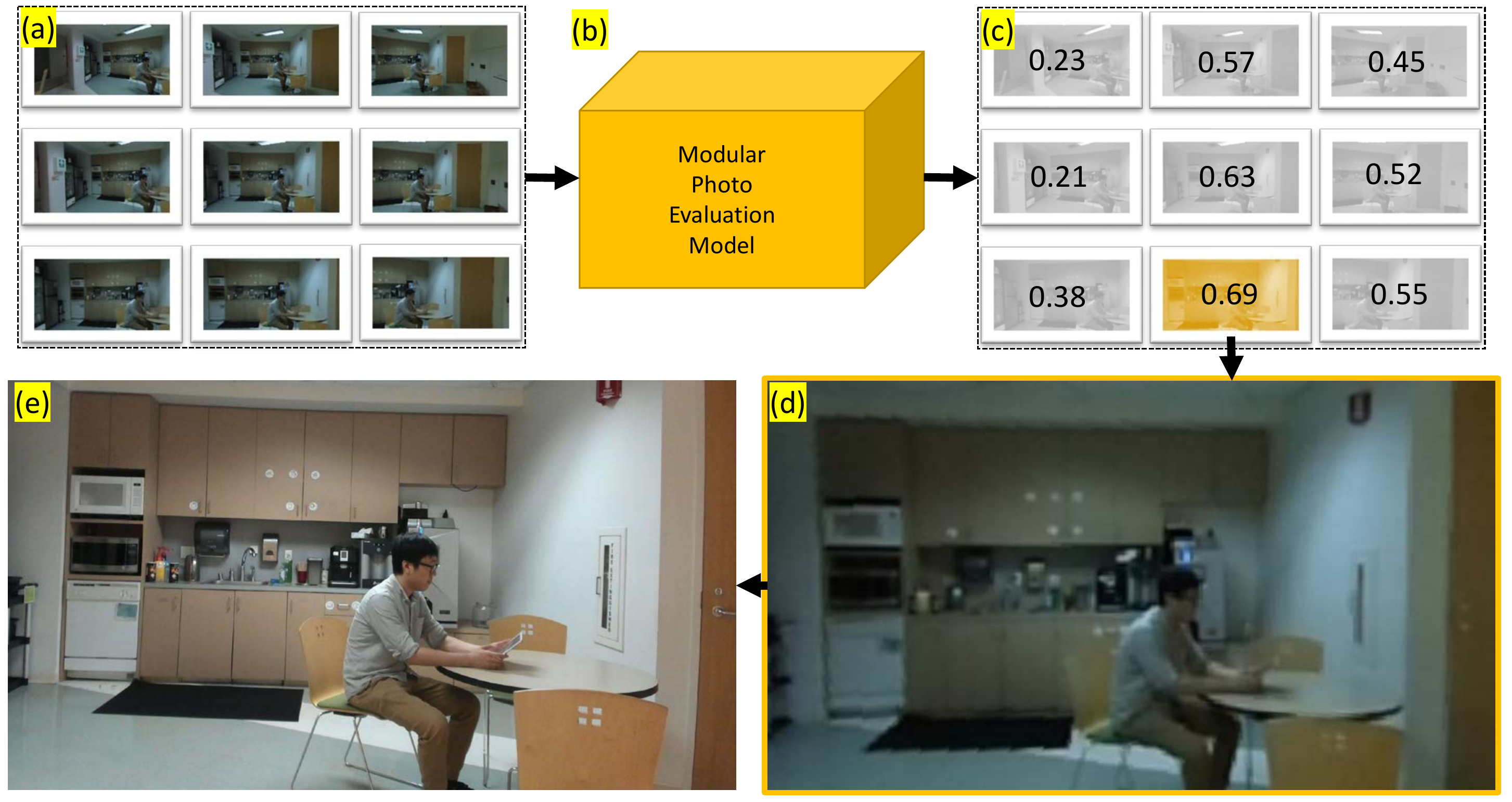}
\caption{Finding the best template. Part (a) presents a collection of candidate templates. The candidate templates are passed through a modular photo evaluation network in (b). Scores are retrieved from the network for each candidate template in (c). The template with the highest score is chosen to be the template in (d). The final photo is composed with the webcam in (e).}
\label{fig:composition}
\end{figure}

The panorama photo processing is performed on the remote PC, and the procedures of finding and using the best template is shown in Figure~\ref{fig:composition}. Candidate templates (Figure~\ref{fig:composition}a) are passed through an off-the-shelf photo evaluation model (Figure~\ref{fig:composition}b). The View Evaluation Net presented in~\cite{Wei2018} is used in the system on the remote PC, which evaluates and scores photos basing on compositions (Figure~\ref{fig:composition}c). The candidate template with the highest score is chosen (Figure~\ref{fig:composition}d). Once the robot finishes Template Matching (Section~\ref{sec:matching}), the final shot is taken with the webcam (Figure~\ref{fig:composition}e). The system is designed to be modular, so that the photo evaluation model can be swapped adaptively with needs.  

\subsubsection{Template Matching}\label{sec:matching}
\begin{figure}[!hbt]
\centering
\includegraphics[width=\linewidth]{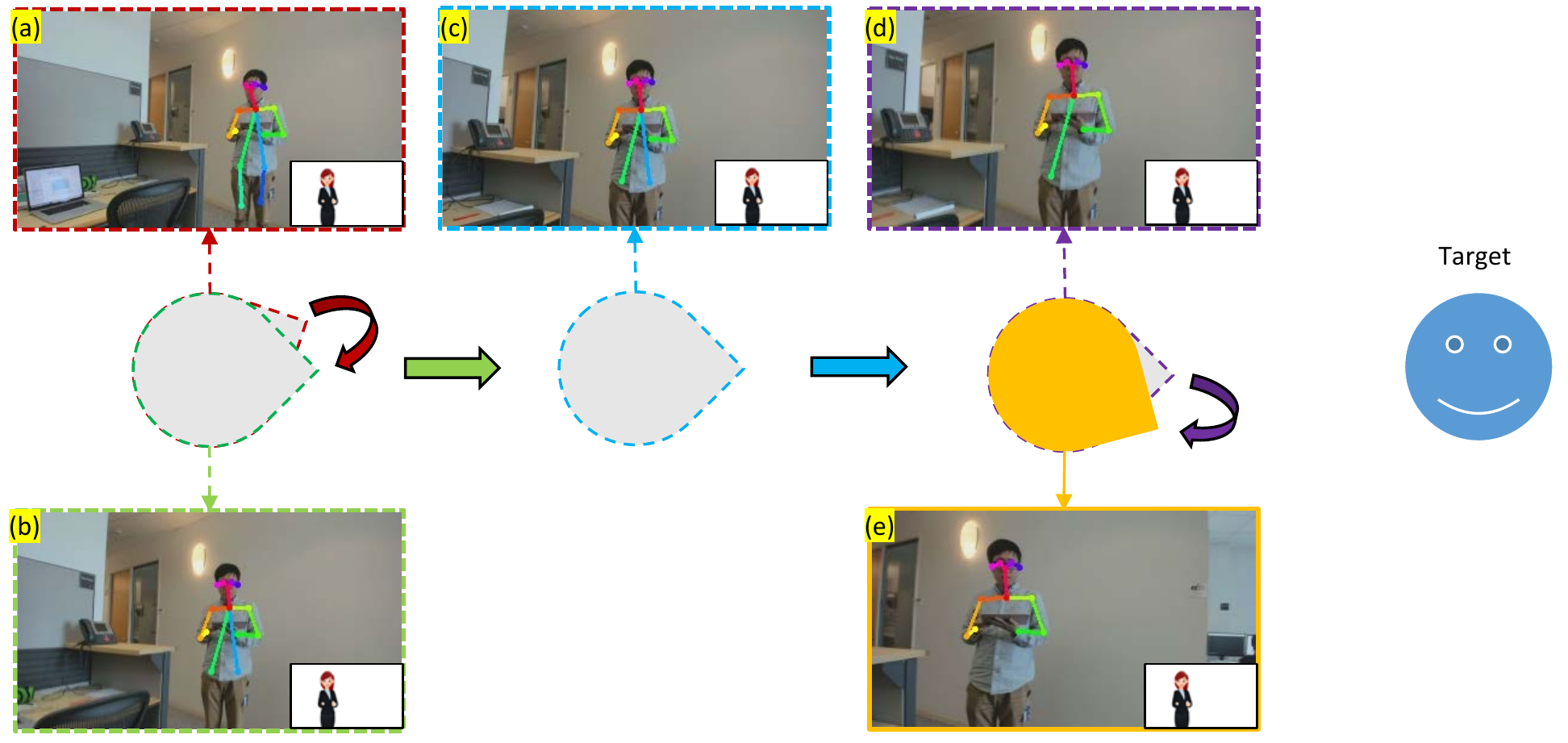}
\caption{A step-by-step template matching example.}
\label{fig:matching}
\end{figure}

The template matching is a process of making the position, size, and pose of the target person in the webcam view as similar as possible to the person in the template by moving the robot to the appropriate location. The similarity can be estimated by comparing the distance of human pose key-points between the current webcam view and the template. OpenPose~\cite{Wei2016} is used to extract the pose key-points and Figure~\ref{fig:matching} demonstrates a simple example of template matching procedures. The template is shown in the bottom right corner in each camera view. Figure~\ref{fig:matching}a is the initial camera view of the robot. The robot first turns right. The target is centered in the camera view as shown in Figure~\ref{fig:matching}b. The robot then moves forward. The target becomes bigger as shown in Figure~\ref{fig:matching}c. The robot moves forward again. The size of the target increases as Figure~\ref{fig:matching}d presents. The robot eventually turns right to reach the end state. The target matches the template with small error below the threshold as shown in Figure~\ref{fig:matching}e. 

The goal of template matching is to get the robot to capture the desired view with the fixed webcam. The input is the visual information from the webcam video stream. The output is a sequence of robot motor actions. The template matching can be simplified as a problem of robot special navigation. Deep Reinforcement Learning (DRL) has been successfully applied to the robot navigation problem in recent studies~\cite{Zhang2017,Mo2018b}. The advantage of using DRL in our framework is that such settings are more adaptive to changes than rule-based and geometric-based solutions. New policies can be simply re-trained with a tuned reward function. 

In typical reinforcement learning settings, the agent receives $(s_{t}, a_{t}, r_{t}, s_{t+1})$ at each time $t$ when interacting with the environment, where $s_{t}$ is the current state, $a_{t}$ is the action the agent makes according to its policy $\pi$, $r_{t}$ is the reward produced by the environment based on $s_{t}$ and $a_{t}$, and $s_{t+1}$ is the next state after transitioning through the environment. The goal is to maximize the expected cumulative return from each state $s_{t}$ in $\mathbb{E}_{\pi}[\sum_{i\geqslant{0}}^{\infty}\gamma^{i}r(s_{t+i}, a_{t+i})]$, where $\gamma$ is the discount factor in $(0,1]$. 

For the template matching, the actions are robot discrete linear and angular velocities. The reward function can be set as an exponential function in Equation~\ref{eqn:reward}.   
\begin{eqnarray}
r &=& \mathrm{e}^{-\alpha||v - v^\prime||},\label{eqn:reward}
\end{eqnarray}
where $v$ and $v^\prime$ represent the current target keypoint vector and the goal template keypoint vector extracted with OpenPose~\cite{Wei2016} respectively, and $\alpha$ is a constant factor (2.5e-03) to adjust the $L^{2}$ distance scale. The DRL training and experiment is discussed in Section~\ref{sec:experiment}. 

\begin{figure}[!hbt]
\centering
\includegraphics[width=0.99\linewidth]{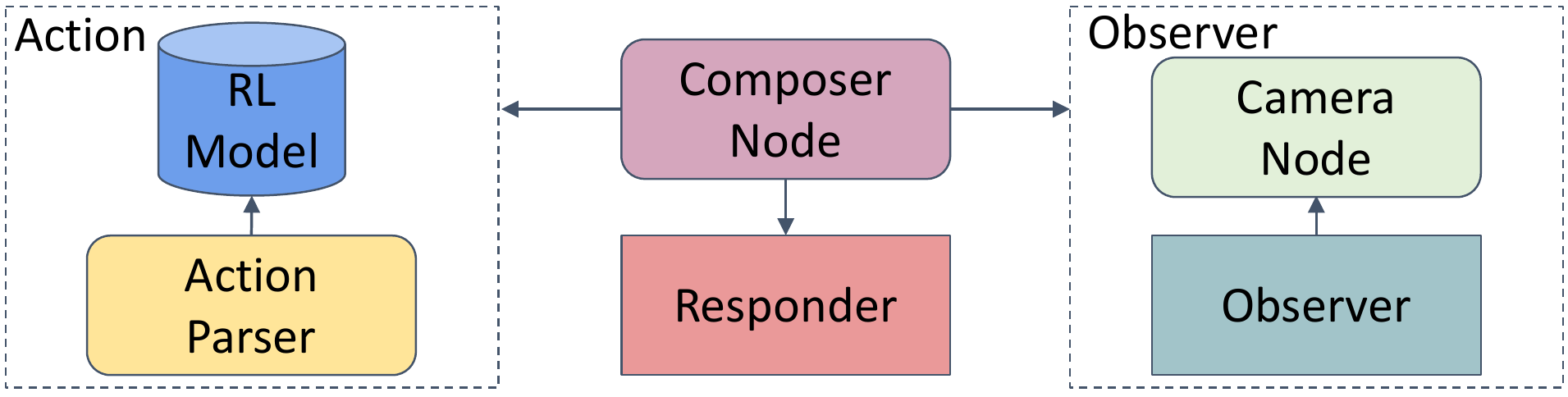}
\caption{The composer module architecture.}
\label{fig:composer_block}
\end{figure}

The architecture of the composer module is demonstrated in Figure~\ref{fig:composer_block}. The \textit{Observer} receives a webcam image from the \textit{Camera Node}, and extracts out human pose key-points as an observation. The \textit{Composer Node} passes the observation to the \textit{Action Parser Node}, which uses a DRL model to predict the next robot action based on the current observation. When an action is decided, the \textit{Action Parser Node} sends it back to the \textit{Composer Node}. The \textit{Responder} parses the robot action into corresponding velocity commands and broadcasts out for execution. The composer module runs on the remote PC.

\subsubsection{The Final Photo Acquisition}
\begin{figure}[!hbt]
\centering
\includegraphics[width=0.6\linewidth]{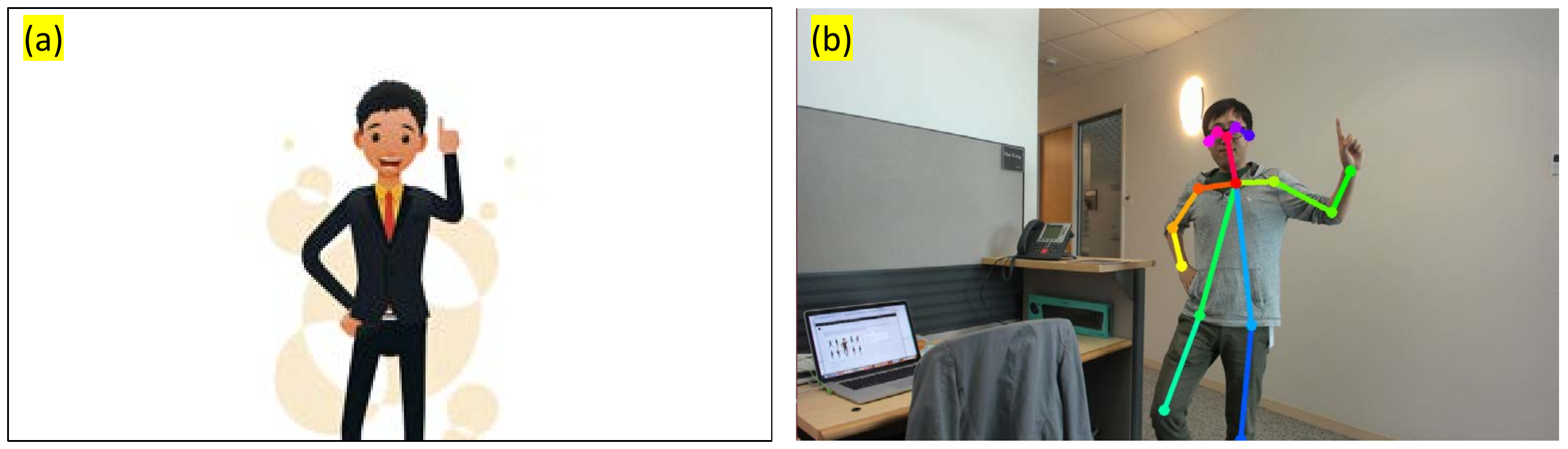}
\caption{Triggering camera shutter with a pose. The template in (a) shows a pre-defined pose template. The photo in (b) shows the target making the pose to trigger the webcam shutter. }
\label{fig:trigger}
\end{figure}

After the current webcam view matches the template, the robot starts capturing photos. The webcam shutter can be triggered with a special pose if a pre-defined pose template is manually selected in the beginning of composing (Figure~\ref{fig:trigger}). The pose triggering is implemented by comparing the similarity of pose key-point \cite{Wei2016} coordinates between the person in template and the person in webcam view. A set of candidate photos are taken during the photo acquisition. The user can manually select the best photo among the candidates, or a best frame selection model such as~\cite{Ren2018} can automatically decide the best photo for the user based on frame or face scores. 

\subsection{Interaction Node}
\label{sec:interaction}
The \textit{Interaction Node} is deployed on an iPad as an iOS application written in Objective-C. It provides the GUI to the user for Human-Robot Interaction. The application has two modes: the \textit{Following Mode}, and the \textit{Photographing Mode}. When the \textit{Following Mode} is activated, the \textit{Tracker Module} is executed. When the \textit{Photographing Mode} is activated, the \textit{Composer Module} is executed. The touch screen takes user input and gives the user on screen graphical output or text-to-speech prompts. The details of the Interaction Node is shown in the complementary video.

\section{Experiments}\label{sec:experiment}
The virtual training environment for Robot Template Matching View Adjustment was set up with synthetic images cropped and processed from panorama photos sampled in real scenes. We trained the view adjustment network with two implementations of DRL. The model was optimized by using action memory and adaptive velocity. We tested the robot photographer in 5 real scenes.

\subsection{Data Preparation}
\begin{figure}[!hbt]
\centering
\includegraphics[width=0.99\linewidth]{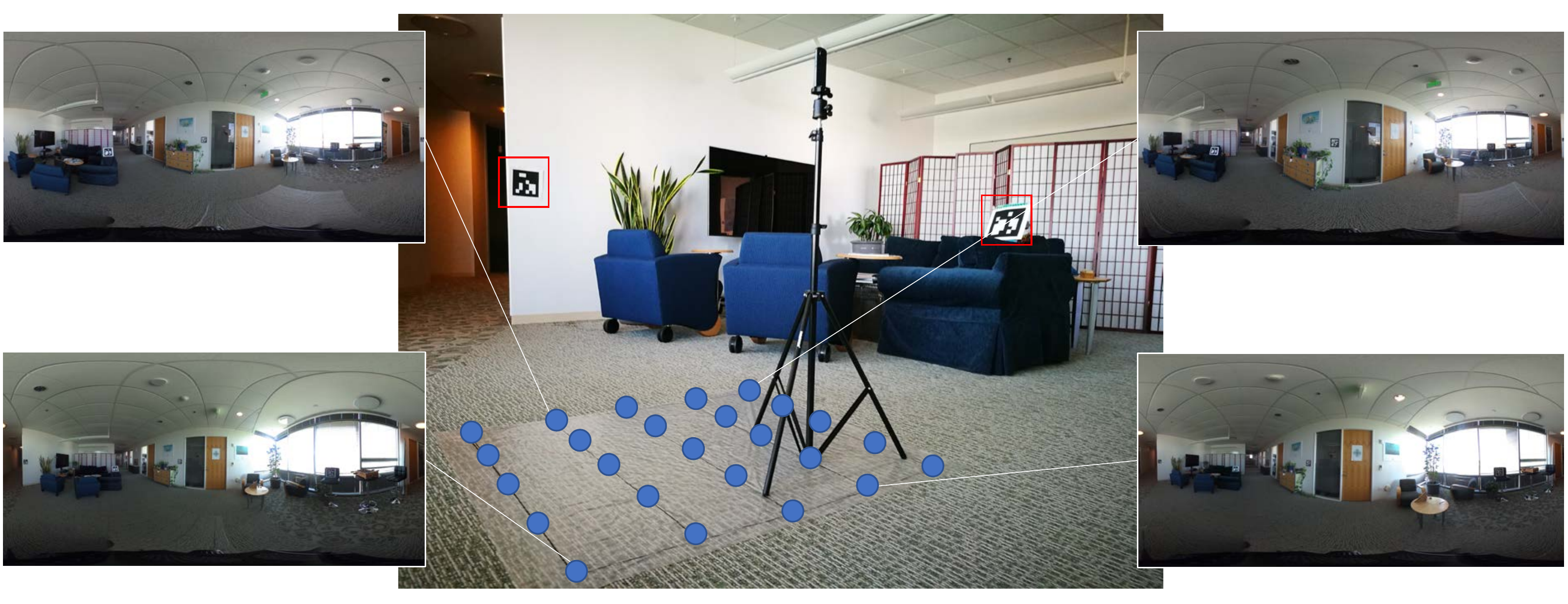}
\caption{Training data collection with a 360 camera on a $5\times5$ grid mat.}
\label{fig:data_collection}
\end{figure}

A Ricoh Theta S 360 camera was used to take panorama photos on a $5\times5$ grid mat (Figure~\ref{fig:data_collection}). One photo was taken at each grid point. The distance between adjacent grid points was $20$ cm. In each scene, 1-5 printed QR codes were placed. A total of 25 photos were collected at each scene; and a total of 15 indoor scenes were selected. 

\begin{figure}[!hbt]
\centering
\includegraphics[width=\linewidth]{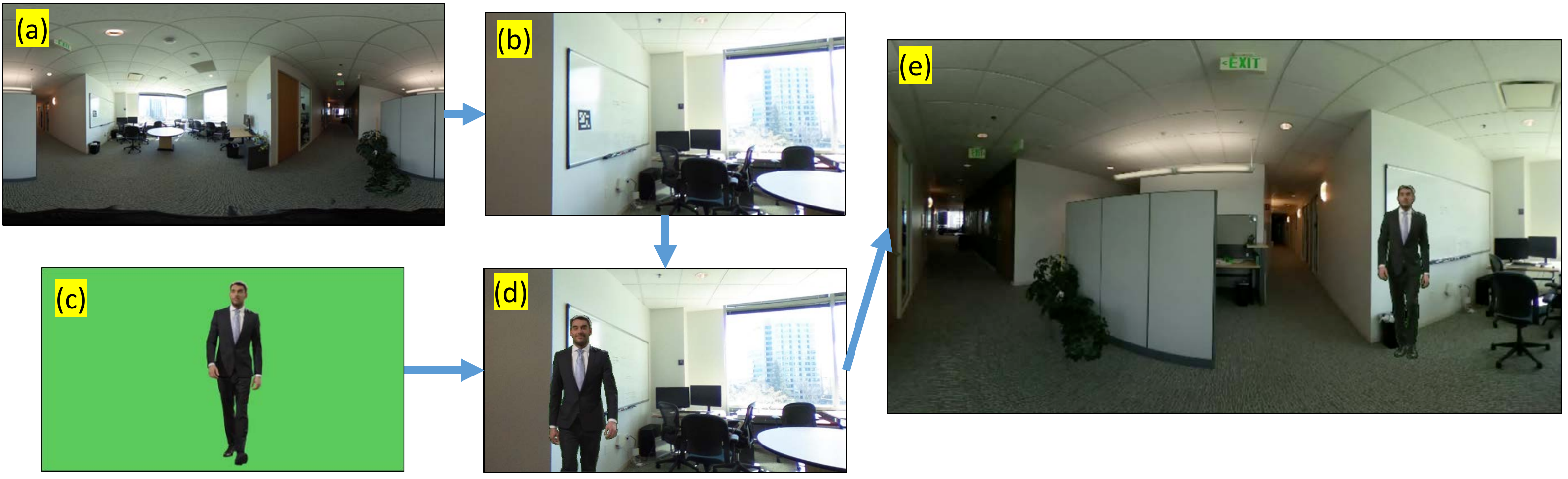}
\caption{Virtual training environment setup.}
\label{fig:training_data}
\end{figure}

Figure~\ref{fig:training_data} indicates the procedures of setting up the virtual training environment. Each $360^{\circ}$ photo (Figure~\ref{fig:training_data}a) was cropped every $15^{\circ}$ into 24 "rotation images" (Figure~\ref{fig:training_data}b). For each scene, there are $600$ images. A person in a short video (Figure~\ref{fig:training_data}c) was cropped out as a collection of frames with minor differences on poses. 30 person videos were processed. One random frame of the cropped person was attached to the location of the QR code for each rotation image (Figure~\ref{fig:training_data}d). The frames of a same person were used for the same QR code in a scene. The scale of the person frame was determined by the size of the QR code in the photo. OpenAI Gym \cite{Brockman2016} virtual environment was set up with $600$ synthesis images for each scene (Figure~\ref{fig:training_data}e). The robot rotation action was simulated with the 24 rotation images from the same $360^{\circ}$ photo. The robot translation action was simulated with the crops of adjacent $360^{\circ}$ photo with the same yaw angle. Nearest neighbour snapping was used in translation simulation. Equation~\ref{eqn:translation} shows the location calculation of a robot translation action.

\begin{eqnarray}
\begin{bmatrix} x\\ y \end{bmatrix} &=& \nint{\begin{bmatrix} x^\prime\\y^\prime\end{bmatrix} + \delta \begin{bmatrix} \sin\theta\\\cos\theta\end{bmatrix}} 
\label{eqn:translation}
\end{eqnarray}
where $[x^\prime,y^\prime]^T$ present the robot initial location coordinate, and $[x, y]^T$ present the robot location coordinate after a translation action. $\theta$ is the yaw angle, and $\delta$ is the signed step scalar which presents the size and direction (forward and backward) of one translation action.

\subsection{Training}
The DRL model was trained with Advantage Actor Critic (A2C) method, which is a synchronous deterministic variant of Asynchronous Advantage Actor Critic (A3C)~\cite{Mnih2016}, and Actor Critic using Kronecker-Factored Trust Region (ACKTR)~\cite{Wu2017} method respectively using a PyTorch implementation \cite{Kostrikov2018}. The environment observation was set to be a vector that contains the selected pose key-point coordinates extracted with OpenPose~\cite{Wei2016}. The robot translation and rotation actions were expressed as one-hot vector array. 

\begin{figure}[!hbt]
\centering
\includegraphics[width=0.99\linewidth]{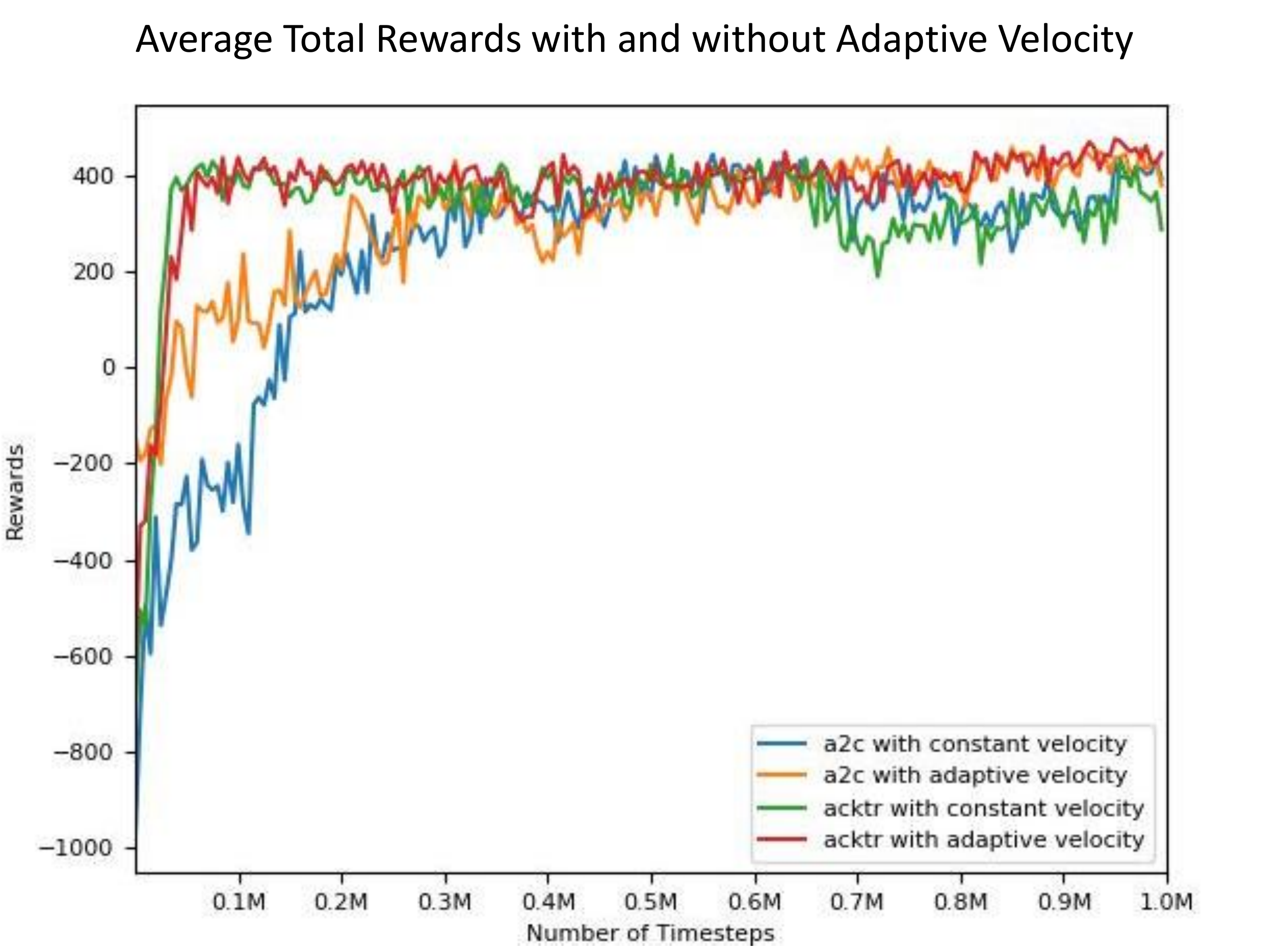}
\caption{Improvements on average total rewards in training. The figure shows a comparison of average total rewards with constant velocity and with 3-level adaptive velocity using A2C and ACKTR. }
\label{fig:reward_vel}
\end{figure}

\begin{figure}[!hbt]
\centering
\includegraphics[width=0.99\linewidth]{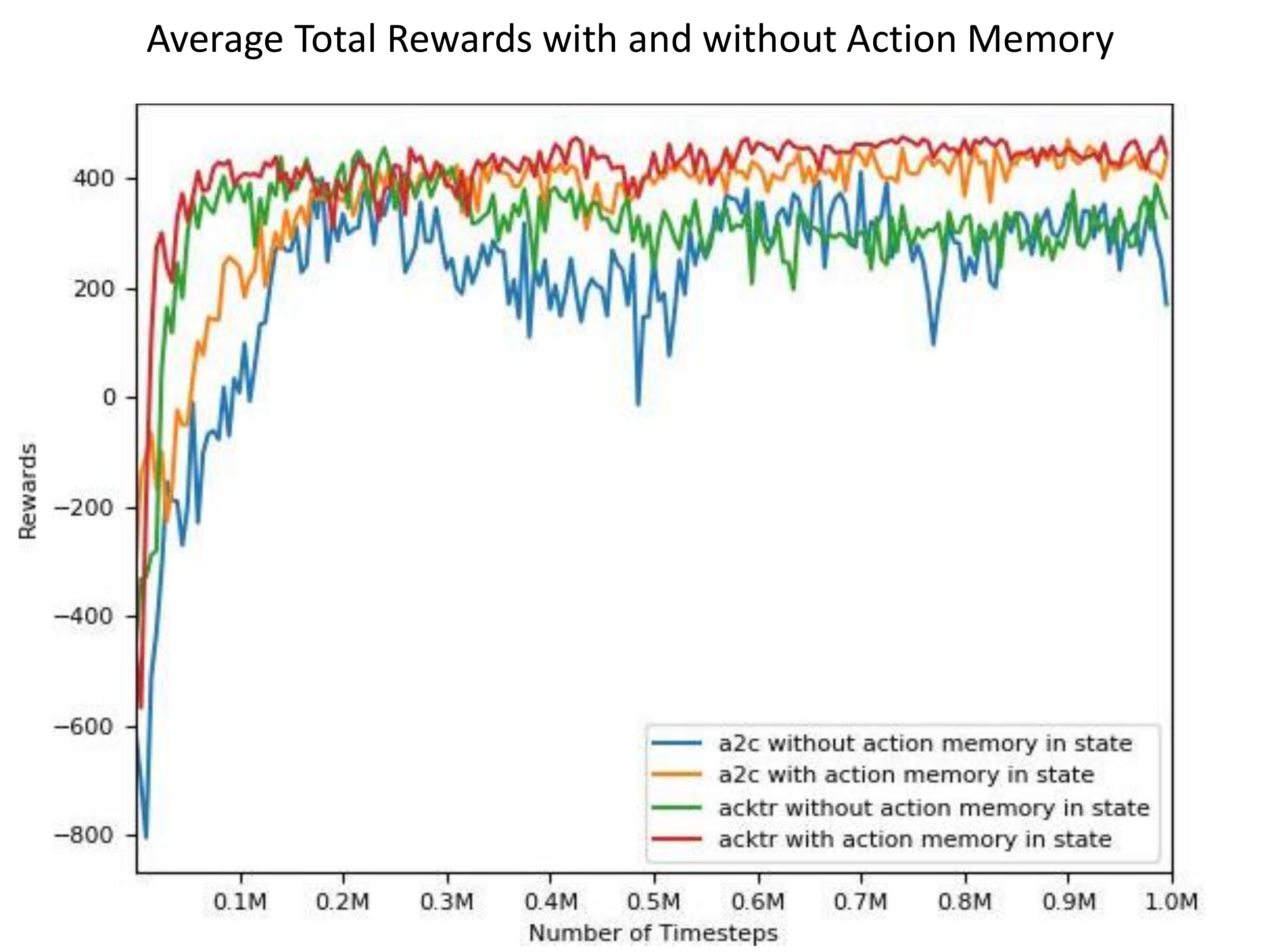}
\caption{Improvements on average total rewards in training. The  figure shows a comparison of average total rewards with and without previous 5 actions as memory using A2C and ACKTR. }
\label{fig:reward_mem}
\end{figure}

In the training, we noticed that the average total rewards of A2C and ACKTR were always on the same level. However, ACKTR tended to converge at an earlier stage than A2C (Figure~\ref{fig:reward_vel} and Figure~\ref{fig:reward_mem}). We experimentally improved the average total rewards by using two methods: (a) adaptive velocity and (b) action memory. The robot actions were reflected as linear and angular velocities in one hot vector array. If we increase the array size by setting different levels of velocities to allow the agent to sample, a slightly higher average total reward can be reached for both A2C and ACKTR as Figure~\ref{fig:reward_vel} shows. Also, if we add previous actions made by the agent to the current observation as memory, both A2C and ACKTR can get higher average total rewards as Figure~\ref{fig:reward_mem} shows. 

\subsection{Testing}
We tested the robot photographer at three indoor scenes, and used it to take 20 photos at each scene (ten using pre-defined template and ten using dynamically generated template). Two of the 60 tests failed to conduct the final capture within 30 actions (1 using pre-defined template, 1 using dynamically generated template) and thus are removed from the test samples. The linear velocity of the robot was set to be no more than $0.15$ m/s; and the angular velocity was set to be no more than $0.5$ rad/s. For dynamically generated template tests, the number of actions has mean $\bar{X}=12.76$, standard deviation $SD=4.67$; and the composing time has $\bar{X}=24.10, SD=8.57$. For pre-defined template tests, the number of action has $\bar{X}=11.20, SD=4.10$; and the composing time has $\bar{X}=22.11, SD=7.47$. The resulting photo evaluation was omitted, and the related user study refers to the off-the-shelf composition model study~\cite{Wei2018}.

\section{Conclusions}
We have developed a novel learning based modular framework for robot photography. The framework allows the robot to take well-composed photographs of a person. The robot photographer has a GUI displayed on an attached iPad that gives voice prompts to the user. Our robot can track the user to a desired location. Then, it starts to adjust its position to form the view in a webcam that matches the best template portrait to capture. The best template is searched by using a modular photo evaluation aesthetic model on cropped images of a panorama photo. The view adjustment is driven by a DRL model basing on template matching. A synthetic virtual environment for navigation training solution was provided. 

The system has several \textit{limitations}. It has limited on-board computation power, so that it relies on a powerful remote PC to run DNN models. The Turtlebot that serves as the basis of our system is relatively simple with few degrees of freedom. 

\textit{Future work} includes testing the solution by using a more complex robot. Also, our system currently supports only a single person portrait. New policies would need to be re-trained to get better support on taking group photos. In future work, we also would like to test different photo evaluation aesthetic models, and extend the work to outdoor scenes. 
\bibliographystyle{./source/IEEEtran}

\begin{thebibliography}{10}
\providecommand{\url}[1]{#1}
\csname url@rmstyle\endcsname
\providecommand{\newblock}{\relax}
\providecommand{\bibinfo}[2]{#2}
\providecommand\BIBentrySTDinterwordspacing{\spaceskip=0pt\relax}
\providecommand\BIBentryALTinterwordstretchfactor{4}
\providecommand\BIBentryALTinterwordspacing{\spaceskip=\fontdimen2\font plus
\BIBentryALTinterwordstretchfactor\fontdimen3\font minus
  \fontdimen4\font\relax}
\providecommand\BIBforeignlanguage[2]{{%
\expandafter\ifx\csname l@#1\endcsname\relax
\typeout{** WARNING: IEEEtran.bst: No hyphenation pattern has been}%
\typeout{** loaded for the language `#1'. Using the pattern for}%
\typeout{** the default language instead.}%
\else
\language=\csname l@#1\endcsname
\fi
#2}}

\bibitem{Grill1990}
T.~Grill and M.~Scanlon, \emph{Photographic composition}.\hskip 1em plus 0.5em
  minus 0.4em\relax Amphoto Books, 1990.

\bibitem{Kim2010}
M.-J. Kim, T.-H. Song, S.-H. Jin, S.-M. Jung, G.-H. Go, K.-H. Kwon, and J.-W.
  Jeon, ``Automatically available photographer robot for controlling
  composition and taking pictures,'' in \emph{IROS 2010}.\hskip 1em plus 0.5em
  minus 0.4em\relax IEEE, 2010, pp. 6010--6015.

\bibitem{Zabarauskas2014}
M.~Zabarauskas and S.~Cameron, ``Luke: An autonomous robot photographer,'' in
  \emph{ICRA 2014}.\hskip 1em plus 0.5em minus 0.4em\relax IEEE, 2014, pp.
  1809--1815.

\bibitem{Chen2014}
J.~Chen and P.~Carr, ``Autonomous camera systems: A survey,'' in
  \emph{Workshops at the 28th AAAI Conference on AI}, 2014, pp. 18--22.

\bibitem{Galvane2013}
Q.~Galvane, M.~Christie, R.~Ronfard, C.-K. Lim, and M.-P. Cani, ``Steering
  behaviors for autonomous cameras,'' in \emph{Proc. of Motion on Games}, ser.
  MIG '13.\hskip 1em plus 0.5em minus 0.4em\relax ACM, 2013, pp. 71:93--71:102.

\bibitem{Galvane2014}
Q.~Galvane, R.~Ronfard, M.~Christie, and N.~Szilas, ``Narrative-driven camera
  control for cinematic replay of computer games,'' in \emph{Proc. of the 7th
  International Conference on Motion in Games}, ser. MIG '14.\hskip 1em plus
  0.5em minus 0.4em\relax ACM, 2014, pp. 109--117.

\bibitem{Galvane2015}
Q.~Galvane, M.~Christie, C.~Lino, and R.~Ronfard, ``Camera-on-rails: Automated
  computation of constrained camera paths,'' in \emph{Proc. of the 8th ACM
  SIGGRAPH Conference on Motion in Games}, ser. MIG '15.\hskip 1em plus 0.5em
  minus 0.4em\relax ACM, 2015, pp. 151--157.

\bibitem{Pinhanez1995}
C.~S. Pinhanez and A.~P. Pentland, \emph{Intelligent studios: Using computer
  vision to control TV cameras}.\hskip 1em plus 0.5em minus 0.4em\relax
  Perceptual Computing Section, Media Laboratory, Massachusetts Institute of
  Technology, 1995.

\bibitem{Pinhanez1997}
C.~S. Pinhanez and A.~F. Bobick, ``Intelligent studios modeling space and
  action to control tv cameras,'' \emph{Applied Artificial Intelligence},
  vol.~11, no.~4, pp. 285--305, 1997.

\bibitem{Byers2003}
Z.~Byers, M.~Dixon, K.~Goodier, C.~M. Grimm, and W.~D. Smart, ``An autonomous
  robot photographer,'' in \emph{IROS 2013}, vol.~3.\hskip 1em plus 0.5em minus
  0.4em\relax IEEE, 2003, pp. 2636--2641.

\bibitem{Smart2003}
W.~D. Smart, C.~M. Grimm, M.~Dixon, and Z.~Byers, ``(not) interacting with a
  robot photographer,'' in \emph{Proc. of the AAAI Spring Symposium on Human
  Interaction with Auto. Syst. in Complex Environments}, 2003.

\bibitem{Byers2004}
Z.~Byers, M.~Dixon, W.~D. Smart, and C.~M. Grimm, ``Say cheese! experiences
  with a robot photographer,'' \emph{AI magazine}, vol.~25, no.~3, p.~37, 2004.

\bibitem{Ahn2006}
H.~Ahn, D.~Kim, J.~Lee, S.~Chi, K.~Kim, J.~Kim, M.~Hahn, and H.~Kim, ``A robot
  photographer with user interactivity,'' in \emph{IROS 2006}.\hskip 1em plus
  0.5em minus 0.4em\relax IEEE, 2006, pp. 5637--5643.

\bibitem{Campbell2005}
J.~{Campbell} and P.~{Pillai}, ``Leveraging limited autonomous mobility to
  frame attractive group photos,'' in \emph{Proc. of the ICRA 2005}, April
  2005, pp. 3396--3401.

\bibitem{Gebhardt2016}
C.~Gebhardt, B.~Hepp, T.~N{\"a}geli, S.~Stev{\v{s}}i{\'c}, and O.~Hilliges,
  ``Airways: Optimization-based planning of quadrotor trajectories according to
  high-level user goals,'' in \emph{Proc. of the 2016 CHI Conference on Human
  Factors in Computing Systems}.\hskip 1em plus 0.5em minus 0.4em\relax ACM,
  2016, pp. 2508--2519.

\bibitem{Richter2016}
C.~Richter, A.~Bry, and N.~Roy, ``Polynomial trajectory planning for aggressive
  quadrotor flight in dense indoor environments,'' in \emph{Robotics
  Research}.\hskip 1em plus 0.5em minus 0.4em\relax Springer, 2016, pp.
  649--666.

\bibitem{Roberts2016}
M.~Roberts and P.~Hanrahan, ``Generating dynamically feasible trajectories for
  quadrotor cameras,'' \emph{ACM Transactions on Graphics (TOG)}, vol.~35,
  no.~4, p.~61, 2016.

\bibitem{Kang2017}
H.~Kang, H.~Li, J.~Zhang, X.~Lu, and B.~Benes, ``Flycam: Multitouch gesture
  controlled drone gimbal photography,'' \emph{IEEE Robotics and Automation
  Letters}, vol.~3, no.~4, pp. 3717--3724, Oct 2018.

\bibitem{Lan2017}
Z.~Lan, M.~Shridhar, D.~Hsu, and S.~Zhao, ``Xpose: Reinventing user interaction
  with flying cameras,'' in \emph{Robotics: Science and Systems}, 2017.

\bibitem{Gadde2011}
R.~Gadde and K.~Karlapalem, ``Aesthetic guideline driven photography by
  robots,'' in \emph{IJCAI Proceedings-International Joint Conference on
  Artificial Intelligence}, vol.~22, no.~3, 2011, p. 2060.

\bibitem{Gooch2001}
B.~Gooch, E.~Reinhard, C.~Moulding, and P.~Shirley, ``Artistic composition for
  image creation,'' in \emph{Rendering Techniques 2001}.\hskip 1em plus 0.5em
  minus 0.4em\relax Springer, 2001, pp. 83--88.

\bibitem{Cavalcanti2006}
C.~Cavalcanti, H.~Gomes, R.~Meireles, and W.~Guerra, ``Towards automating
  photographic composition of people,'' in \emph{Proc. of the IASTED
  International Conference on Visualization, Imaging, and Image Processing},
  2006, pp. 25--30.

\bibitem{Banerjee2007}
S.~Banerjee and B.~L. Evans, ``In-camera automation of photographic composition
  rules,'' \emph{IEEE Transactions on Image Processing}, vol.~16, no.~7, pp.
  1807--1820, 2007.

\bibitem{Datta2006}
R.~Datta, D.~Joshi, J.~Li, and J.~Z. Wang, ``Studying aesthetics in
  photographic images using a computational approach,'' in \emph{ECCV
  2006}.\hskip 1em plus 0.5em minus 0.4em\relax Springer, 2006, pp. 288--301.

\bibitem{Dhar2011}
S.~Dhar, V.~Ordonez, and T.~L. Berg, ``High level describable attributes for
  predicting aesthetics and interestingness,'' in \emph{CVPR 2011}.\hskip 1em
  plus 0.5em minus 0.4em\relax IEEE, 2011, pp. 1657--1664.

\bibitem{Ke2006}
Y.~Ke, X.~Tang, and F.~Jing, ``The design of high-level features for photo
  quality assessment,'' in \emph{CVPR 2006}, vol.~1.\hskip 1em plus 0.5em minus
  0.4em\relax IEEE, 2006, pp. 419--426.

\bibitem{Luo2011}
W.~Luo, X.~Wang, and X.~Tang, ``Content-based photo quality assessment,'' in
  \emph{ICCV 2011}.\hskip 1em plus 0.5em minus 0.4em\relax IEEE, 2011, pp.
  2206--2213.

\bibitem{Nishiyama2011}
M.~Nishiyama, T.~Okabe, I.~Sato, and Y.~Sato, ``Aesthetic quality
  classification of photographs based on color harmony,'' in \emph{CVPR
  2011}.\hskip 1em plus 0.5em minus 0.4em\relax IEEE, 2011, pp. 33--40.

\bibitem{Murray2012}
N.~Murray, L.~Marchesotti, and F.~Perronnin, ``Ava: A large-scale database for
  aesthetic visual analysis,'' in \emph{CVPR 2012}.\hskip 1em plus 0.5em minus
  0.4em\relax IEEE, 2012, pp. 2408--2415.

\bibitem{Kong2016}
S.~Kong, X.~Shen, Z.~Lin, R.~Mech, and C.~Fowlkes, ``Photo aesthetics ranking
  network with attributes and content adaptation,'' in \emph{ECCV 2016}.\hskip
  1em plus 0.5em minus 0.4em\relax Springer, 2016, pp. 662--679.

\bibitem{Wei2018}
Z.~Wei, J.~Zhang, X.~Shen, Z.~Lin, R.~Mech, M.~Hoai, and D.~Samaras, ``Good
  view hunting: Learning photo composition from dense view pairs,'' in
  \emph{CVPR 2018}.\hskip 1em plus 0.5em minus 0.4em\relax IEEE, 2018, pp.
  5437--5446.

\bibitem{Lu2015}
X.~Lu, Z.~Lin, X.~Shen, R.~Mech, and J.~Z. Wang, ``Deep multi-patch aggregation
  network for image style, aesthetics, and quality estimation,'' in \emph{Proc.
  ICCV 2015}.\hskip 1em plus 0.5em minus 0.4em\relax IEEE, 2015, pp. 990--998.

\bibitem{Lu2014}
X.~Lu, Z.~Lin, H.~Jin, J.~Yang, and J.~Z. Wang, ``Rapid: Rating pictorial
  aesthetics using deep learning,'' in \emph{Proceedings of the 22nd ACM
  international conference on Multimedia}.\hskip 1em plus 0.5em minus
  0.4em\relax ACM, 2014, pp. 457--466.

\bibitem{Marchesotti2015}
L.~Marchesotti, N.~Murray, and F.~Perronnin, ``Discovering beautiful attributes
  for aesthetic image analysis,'' \emph{International journal of computer
  vision}, vol. 113, no.~3, pp. 246--266, 2015.

\bibitem{Mai2016}
L.~Mai, H.~Jin, and F.~Liu, ``Composition-preserving deep photo aesthetics
  assessment,'' in \emph{Proc. of CVPR 2016}.\hskip 1em plus 0.5em minus
  0.4em\relax IEEE, 2016, pp. 497--506.

\bibitem{Kang2014}
L.~Kang, P.~Ye, Y.~Li, and D.~Doermann, ``Convolutional neural networks for
  no-reference image quality assessment,'' in \emph{Proc. of CVPR 2014}.\hskip
  1em plus 0.5em minus 0.4em\relax IEEE, 2014, pp. 1733--1740.

\bibitem{Mnih2015}
V.~Mnih, K.~Kavukcuoglu, D.~Silver, A.~A. Rusu, J.~Veness, M.~G. Bellemare,
  A.~Graves, M.~Riedmiller, A.~K. Fidjeland, G.~Ostrovski, \emph{et~al.},
  ``Human-level control through deep reinforcement learning,'' \emph{Nature},
  vol. 518, no. 7540, p. 529, 2015.

\bibitem{Silver2016}
D.~Silver, A.~Huang, C.~J. Maddison, A.~Guez, L.~Sifre, G.~Van Den~Driessche,
  J.~Schrittwieser, I.~Antonoglou, V.~Panneershelvam, M.~Lanctot,
  \emph{et~al.}, ``Mastering the game of go with deep neural networks and tree
  search,'' \emph{nature}, vol. 529, no. 7587, p. 484, 2016.

\bibitem{Peng2017}
X.~B. Peng, G.~Berseth, K.~Yin, and M.~van~de Panne, ``Deeploco: Dynamic
  locomotion skills using hierarchical deep reinforcement learning,'' \emph{ACM
  Transactions on Graphics (Proc. SIGGRAPH 2017)}, vol.~36, no.~4, 2017.

\bibitem{Hodgins2017}
J.~H. Libin~Liu, ``Learning basketball dribbling skills using trajectory
  optimization and deep reinforcement learning,'' \emph{ACM Transactions on
  Graphics}, vol.~37, no.~4, August 2018.

\bibitem{Gu2017}
S.~Gu, E.~Holly, T.~Lillicrap, and S.~Levine, ``Deep reinforcement learning for
  robotic manipulation with asynchronous off-policy updates,'' in \emph{2017
  IEEE International Conference on Robotics and Automation (ICRA)}.\hskip 1em
  plus 0.5em minus 0.4em\relax IEEE, 2017, pp. 3389--3396.

\bibitem{Hwangbo2017}
J.~Hwangbo, I.~Sa, R.~Siegwart, and M.~Hutter, ``Control of a quadrotor with
  reinforcement learning,'' \emph{IEEE Robotics and Automation Letters},
  vol.~2, no.~4, pp. 2096--2103, 2017.

\bibitem{Levine2016}
S.~Levine, C.~Finn, T.~Darrell, and P.~Abbeel, ``End-to-end training of deep
  visuomotor policies,'' \emph{The Journal of Machine Learning Research},
  vol.~17, no.~1, pp. 1334--1373, 2016.

\bibitem{Levine2018}
S.~Levine, P.~Pastor, A.~Krizhevsky, J.~Ibarz, and D.~Quillen, ``Learning
  hand-eye coordination for robotic grasping with deep learning and large-scale
  data collection,'' \emph{The International Journal of Robotics Research},
  vol.~37, no. 4-5, pp. 421--436, 2018.

\bibitem{Mirowski2016}
\BIBentryALTinterwordspacing
P.~Mirowski, R.~Pascanu, F.~Viola, H.~Soyer, A.~J. Ballard, A.~Banino,
  M.~Denil, R.~Goroshin, L.~Sifre, K.~Kavukcuoglu, D.~Kumaran, and R.~Hadsell,
  ``Learning to navigate in complex environments,'' \emph{CoRR}, vol.
  abs/1611.03673, 2016. [Online]. Available:
  \url{http://arxiv.org/abs/1611.03673}
\BIBentrySTDinterwordspacing

\bibitem{Zhu2017}
Y.~{Zhu}, R.~{Mottaghi}, E.~{Kolve}, J.~J. {Lim}, A.~{Gupta}, L.~{Fei-Fei}, and
  A.~{Farhadi}, ``Target-driven visual navigation in indoor scenes using deep
  reinforcement learning,'' in \emph{2017 IEEE International Conference on
  Robotics and Automation (ICRA)}, May 2017, pp. 3357--3364.

\bibitem{Quigley2009}
M.~Quigley, K.~Conley, B.~P. Gerkey, J.~Faust, T.~Foote, J.~Leibs, R.~Wheeler,
  and A.~Y. Ng, ``Ros: an open-source robot operating system,'' in \emph{ICRA
  Workshop on Open Source Software}, 2009.

\bibitem{Yujin2018}
I.~Kostrikov, ``Pytorch implementations of reinforcement learning algorithms,''
  \url{https://github.com/ikostrikov/pytorch-a2c-ppo-acktr}, 2018.

\bibitem{Redmon2016}
J.~{Redmon} and A.~{Farhadi}, ``Yolo9000: Better, faster, stronger,'' in
  \emph{CVPR 2017}, July 2017, pp. 6517--6525.

\bibitem{Hermans2017}
A.~Hermans, L.~Beyer, and B.~Leibe, ``In defense of the triplet loss for person
  re-identification,'' \emph{arXiv preprint arXiv:1703.07737}, 2017.

\bibitem{Xiao2017}
T.~Xiao, S.~Li, B.~Wang, L.~Lin, and X.~Wang, ``Joint detection and
  identification feature learning for person search,'' in \emph{CVPR
  2017}.\hskip 1em plus 0.5em minus 0.4em\relax IEEE, 2017, pp. 3376--3385.

\bibitem{Mo2018}
K.~Mo, ``Panorama image viewer and cropper,''
  \url{https://github.com/daerduoCarey/PanoramaImageViewer}, 2018.

\bibitem{Wei2016}
S.-E. Wei, V.~Ramakrishna, T.~Kanade, and Y.~Sheikh, ``Convolutional pose
  machines,'' in \emph{CVPR}, 2016.

\bibitem{Zhang2017}
J.~Zhang, J.~T. Springenberg, J.~Boedecker, and W.~Burgard, ``Deep
  reinforcement learning with successor features for navigation across similar
  environments,'' in \emph{IROS 2017}, Sept 2017, pp. 2371--2378.

\bibitem{Mo2018b}
K.~Mo, H.~Li, Z.~Lin, and J.-Y. Lee, ``The adobeindoornav dataset: Towards deep
  reinforcement learning based real-world indoor robot visual navigation,''
  \emph{arXiv preprint arXiv:1802.08824}, 2018.

\bibitem{Ren2018}
J.~Ren, X.~Shen, Z.~Lin, and R.~Měch, ``{Best Frame Selection in a Short
  Video},'' Adobe Research, Tech. Rep., 2018.

\bibitem{Brockman2016}
\BIBentryALTinterwordspacing
G.~Brockman, V.~Cheung, L.~Pettersson, J.~Schneider, J.~Schulman, J.~Tang, and
  W.~Zaremba, ``Openai gym,'' \emph{CoRR}, vol. abs/1606.01540, 2016. [Online].
  Available: \url{http://arxiv.org/abs/1606.01540}
\BIBentrySTDinterwordspacing

\bibitem{Mnih2016}
V.~Mnih, A.~P. Badia, M.~Mirza, A.~Graves, T.~Lillicrap, T.~Harley, D.~Silver,
  and K.~Kavukcuoglu, ``Asynchronous methods for deep reinforcement learning,''
  in \emph{International conference on machine learning}, 2016, pp. 1928--1937.

\bibitem{Wu2017}
Y.~Wu, E.~Mansimov, R.~B. Grosse, S.~Liao, and J.~Ba, ``Scalable trust-region
  method for deep reinforcement learning using kronecker-factored
  approximation,'' in \emph{Advances in Neural Information Processing Systems
  30}.\hskip 1em plus 0.5em minus 0.4em\relax Curran Associates, Inc., 2017,
  pp. 5279--5288.

\bibitem{Kostrikov2018}
Yujin, ``Yocs velocity smoother,''
  \url{https://github.com/yujinrobot/yujin\_ocs}, 2018.

\end{thebibliography}


\end{document}